%% file: tacl2018v2-template.tex
\documentclass[11pt,a4paper]{article}
\usepackage{times,latexsym}
\usepackage{url}
\usepackage[T1]{fontenc}

\usepackage{enumitem}
\usepackage{times}

\usepackage{soul}
\usepackage[hidelinks]{hyperref}
\usepackage[utf8]{inputenc}
\usepackage[small]{caption}
\usepackage{graphicx}
\usepackage{amsmath}
\usepackage{booktabs}
\usepackage{multirow}
\usepackage{amssymb}
\usepackage{footnote}
\usepackage{rotating}
\usepackage{makecell}
\usepackage{tcolorbox}

\usepackage{fixmath}
\usepackage{lipsum}                     % Dummytext
\usepackage{xargs}                      % Use more than one optional parameter in a new commands

\usepackage[colorinlistoftodos,prependcaption,textsize=small]{todonotes}
\presetkeys{todonotes}{inline}{}
\usepackage[symbol]{footmisc}
\usepackage{tablefootnote}

\let\oldbibliography\thebibliography
\renewcommand{\thebibliography}[1]{%
  \oldbibliography{#1}%
  \setlength{\itemsep}{0pt}%
}

\usepackage{bigfoot}

\DeclareNewFootnote{A}

\MakePerPage{footnoteA}
\DeclareNewFootnote{B}

\usepackage[acceptedWithA]{tacl2018v2}
\usepackage{xspace,mfirstuc,tabulary}

\newif\iftaclinstructions
\taclinstructionsfalse % AUTHORS: do NOT set this to true
\iftaclinstructions
\renewcommand{\confidential}{}
\renewcommand{\anonsubtext}{(No author info supplied here, for consistency with
TACL-submission anonymization requirements)}
\newcommand{\instr}
\fi

\iftaclpubformat % this "if" is set by the choice of options

\else

\fi

\begin{document}

\title{Compressing Large-Scale Transformer-Based Models: \\ A Case Study on BERT}

% Author information does not appear in the pdf unless the "acceptedWithA" option is given
% See tacl2018v2.sty for other ways to format author information
\author{
 Prakhar Ganesh$^{1*}$, Yao Chen$^1$\thanks{$^*$Both authors contributed equally to this research.} , Xin Lou$^1$, Mohammad Ali Khan$^1$, \\ 
 {\bf Yin Yang$^2$, Hassan Sajjad$^3$, Preslav Nakov$^3$, Deming Chen$^4$, Marianne Winslett$^4$} \\
 $^1$Advanced Digital Sciences Center, Singapore \\
 $^2$College of Science and Engineering, Hamad Bin Khalifa University, Qatar \\
 $^3$Qatar Computing Research Institute, Hamad Bin Khalifa University, Qatar \\
 $^4$University of Illinois at Urbana-Champaign, USA \\
 {\sf \{prakhar.g, yao.chen, lou.xin, mohammad.k\}@adsc-create.edu.sg,} \\
 {\sf \{yyang, hsajjad, pnakov\}@hbku.edu.qa, \{dchen, winslett\}@illinois.edu} \\
}
% \footnote{Both authors make equal contribution to this research.}
% \date{}

\maketitle
\begin{abstract}
Pre-trained Transformer-based models have achieved state-of-the-art performance for various Natural Language Processing (NLP) tasks. However, these models often have billions of parameters, and, thus, are too resource-hungry and computation-intensive to suit low-capability devices or applications with strict latency requirements. One potential remedy for this is model compression, which has attracted a lot of research attention. Here, we summarize the research in compressing Transformers, focusing on the especially popular BERT model. In particular, we survey the state of the art in compression for BERT, we clarify the current best practices for compressing large-scale Transformer models, and we provide insights into the workings of various methods. Our categorization and analysis also shed light on promising future research directions for achieving lightweight, accurate, and generic NLP models.
\end{abstract}

\input{sections/01-Introduction.tex}
\input{sections/02-Breakdown_Analysis.tex}
\input{sections/03-Compression_Methods.tex}
\input{sections/04-Evaluation.tex}
\input{sections/05-Open_Issues.tex}

\section*{Acknowledgement}

This publication was made possible by NPRP grant NPRP10-0208-170408 from the Qatar National Research Fund (a member of Qatar Foundation). The findings herein reflect the work, and are solely the responsibility of the authors.
This work is also partially supported by the National Research Foundation, Prime Minister's Office, Singapore under its Campus for Research Excellence and Technological Enterprise (CREATE) programme.

\bibliography{tacl2018}
\bibliographystyle{acl_natbib}

\end{document}

%% file: sections/01-Introduction.tex
\section{Introduction}\label{sec:intro}

\begin{figure}[]
\centering
\includegraphics[width = 0.46\textwidth]{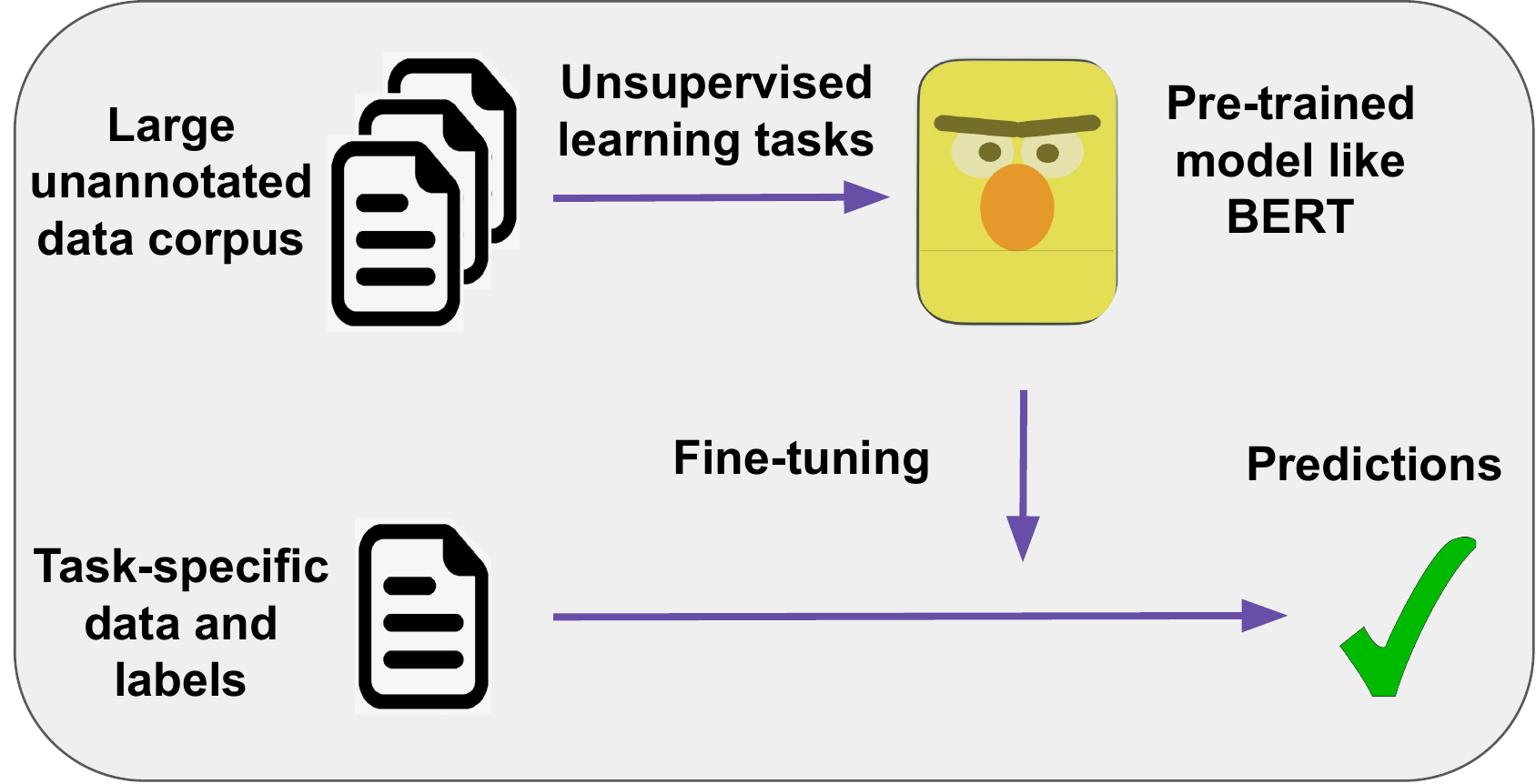}
\caption{Pre-training large-scale models.}
\label{fig:language_models}
\end{figure}

Sentiment analysis, paraphrase detection, machine reading comprehension, question answering, text summarization: all these Natural Language Processing (NLP) tasks benefit from pre-training a large-scale generic model on an enormous corpus such as a Wikipedia dump and/or a book collection, and then fine-tuning for specific downstream tasks, as shown in Fig.~\ref{fig:language_models}. Earlier solutions following this methodology used recurrent neural networks (RNNs) as the generic base model, \textit{e.g.},~ULMFiT \cite{howard2018universal} and ELMo \cite{peters2018deep}.  More recent methods mostly use the Transformer architecture \cite{vaswani2017attention}, which relies heavily on the attention mechanism, \textit{e.g.},~BERT \cite{devlin2019bert}, GPT-2 \cite{radford2019language}, XLNet \cite{yang2019xlnet}, MegatronLM \cite{shoeybi2019megatron}, Turing-NLG \cite{blog_2020}, T5 \cite{raffel2019exploring}, and GPT-3 \cite{brown2020language}.

These Transformers are powerful, \textit{e.g.}, BERT, when first released, improved the state of the art for eleven NLP tasks by sizable margins \cite{devlin2019bert}. However, Transformers are also bulky and resource-hungry: for instance, GPT-3 \cite{brown2020language}, a recent large-scale Transformer, has over 175 billion parameters. Models of this size incur high memory consumption, computational overhead, and energy. 
The problem is exacerbated when we consider devices with lower capacity, \textit{e.g.},~smartphones, and applications with strict latency constraints, \textit{e.g.},~interactive chatbots.

To put things in perspective, a single training run for GPT-3 \cite{brown2020language}, one of the most powerful and heaviest Transformer-based models, trained on a total of 300 billion tokens, costs well above 12 million USD \cite{floridi2020gpt}. Moreover, fine-tuning or even inference with such a model on a downstream task cannot be done on a GPU with 32GB memory, which is the capacity of Tesla V100, one of the most advanced data center GPUs. Instead it requires access to high-performance GPU or multi-core CPU clusters, which often means a need to access cloud computing with high computation density, such as the Google Cloud Platform (GCP), Microsoft Azure, Amazon Web Services (AWS), \textit{etc.}, and results in a high monetary cost \cite{floridi2020gpt}.

One way to address this problem is through model compression, an intricate part of deep learning that has attracted attention from both researchers and practitioners.
A recent study by \citet{li2020train} highlights the importance of first training over-parameterized models and then compressing them, instead of directly training smaller models, to reduce the performance errors.
Although most methods in model compression were originally proposed for convolutional neural networks (CNNs), \textit{e.g.}, pruning, quantization, knowledge distillation, \textit{etc.}~\cite{cheng2017survey}, many ideas are directly applicable to Transformers.
There are also methods designed specifically for Transformers, \textit{e.g.},~attention head pruning, attention decomposition, replacing Transformer blocks with an RNN or a CNN, \textit{etc.} (discussed in Section \ref{sec:compression}). Unlike CNNs, a Transformer model has a relatively complex architecture consisting of multiple parts such as embedding layers, self-attention, and feed-forward layers (details introduced in Section~\ref{sec:breakdown}). Thus, the effectiveness of different compression methods can vary when applied to different parts of a Transformer model.

Several recent surveys have focused on pre-trained representations and large-scale Transformer-based models \textit{e.g.},~\citet{qiu2020pre,rogers2020primer,wang2019survey}. However, to the best of our knowledge, no comprehensive, systematic study has compared the effectiveness of different model compression techniques on Transformer-based large-scale NLP models, even though a variety of approaches for compressing such models have been proposed.
\textbf{Motivated by this, here we offer a thorough and in-depth comparative study on compressing Transformer-based NLP models, with a special focus on the widely used BERT} \cite{devlin2019bert}.  Although the compression methods discussed here can be extended to Transformer-based decoders and multilingual Transformer models, we restrict our discussion to BERT in order to provide detailed insights into various methods.

Our study is timely, since (\emph{i})~the use of Transformer-based BERT-like models has grown dramatically, as demonstrated by current leaders of various NLP tasks such as language understanding \cite{wang2018glue}, machine reading comprehension \cite{rajpurkar2016squad, rajpurkar2018know}, machine translation \cite{machavcek2014results}, summarization \cite{narayan2018don}, \textit{etc.}; (\emph{ii})~many researchers are left behind as they do not have expensive GPUs (or a multi-GPU setup) with a large amount of GPU memory, and thus cannot fine-tune and use the large BERT model for relevant downstream tasks; and (\emph{iii})~AI-powered devices such as smartphones would benefit tremendously from an on-board BERT-like model, but do not have the capability to run it. In addition to summarizing existing techniques and best practices for BERT compression, we point out several promising future directions of research for compressing large-scale Transformer-based models.

%% file: sections/02-Breakdown_Analysis.tex
\section{Breakdown \& Analysis of BERT}\label{sec:breakdown}

Bidirectional Encoder Representations from Transformers, or BERT \cite{devlin2019bert}, is a Transformer model~\cite{vaswani2017attention} pre-trained on large corpora from Wikipedia and the Bookcorpus \cite{zhu2015aligning} using two training objectives: (\emph{i})~Masked Language Model (MLM), which helps it learn the context in a sentence, and (\emph{ii})~Next Sentence Prediction (NSP), from which it learns the relationship between two sentences. Subsequent Transformers have further improved the training objective in various ways \cite{lan2019albert,liu2019roberta}.
In the following, we focus on the original BERT model.

BERT decomposes its input sentence(s) into WordPiece tokens \cite{wu2016google}.
Specifically, WordPiece tokenization helps improve the representation of the input vocabulary and reduce its size, by segmenting complex words into subwords. These subwords can even form new words not seen in the training samples, thus making the model more robust to out-of-vocabulary (OOV) words.
A classification token ([CLS]) is inserted before the input, and the output corresponding to this token is used for classification tasks. For sentence pair tasks, the two sentences are packed together by inserting a separator token ([SEP]) between them.

BERT represents each WordPiece token with three vectors, namely its token, segment, and position embeddings. These embeddings are summed together and then passed through the main body of the model, \textit{i.e.},~the Transformer backbone, which produces the output representations that are fed into the final, application-dependent layer, \textit{e.g.},~a classifier for sentiment analysis.

\begin{figure}
\centering
\includegraphics[width = 0.45\textwidth]{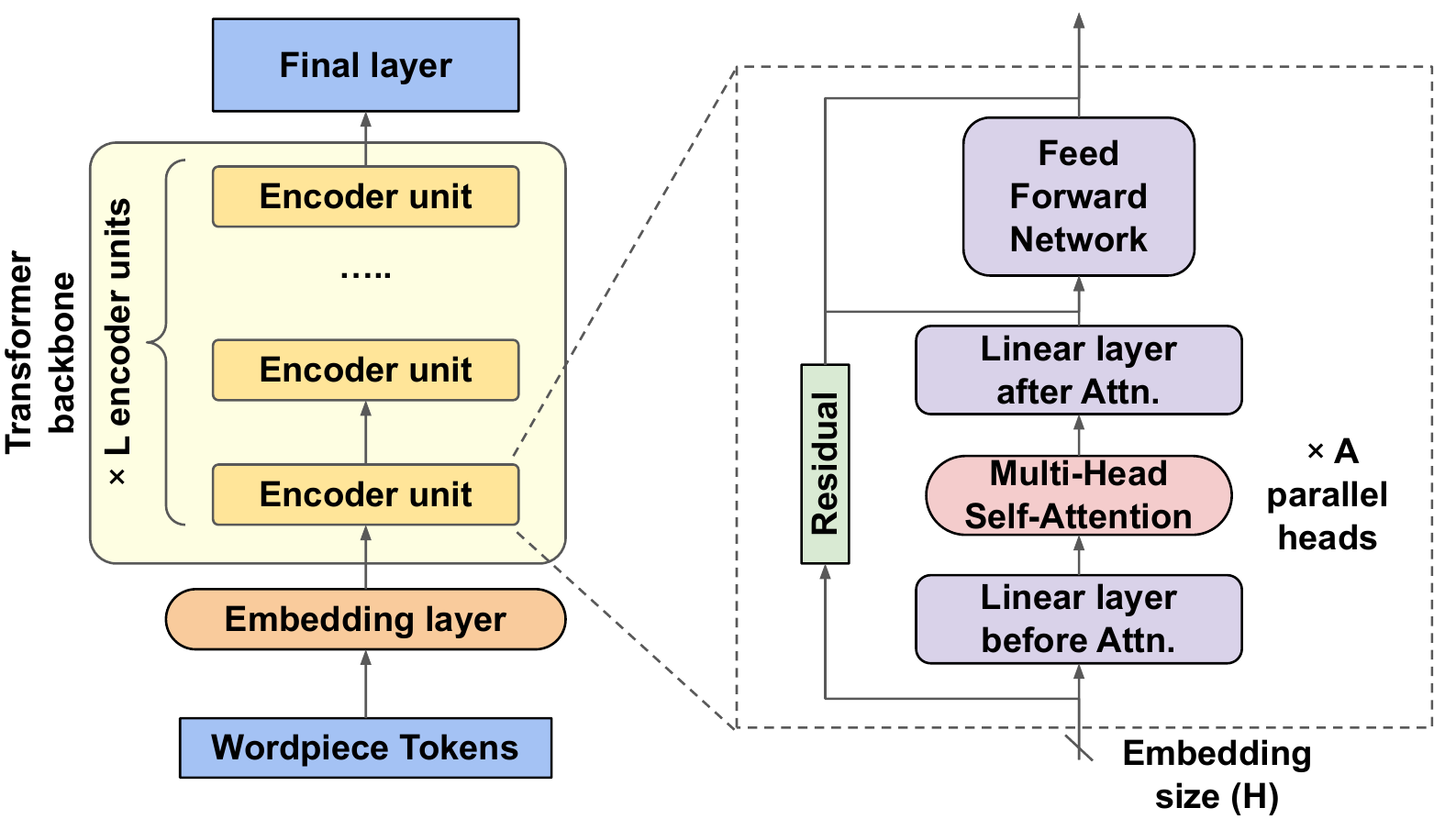}
\caption{BERT model flowchart.}
\label{fig:bert_breakdown}
\end{figure}

\begin{figure*}
\centering
\includegraphics[width = 0.75\textwidth]{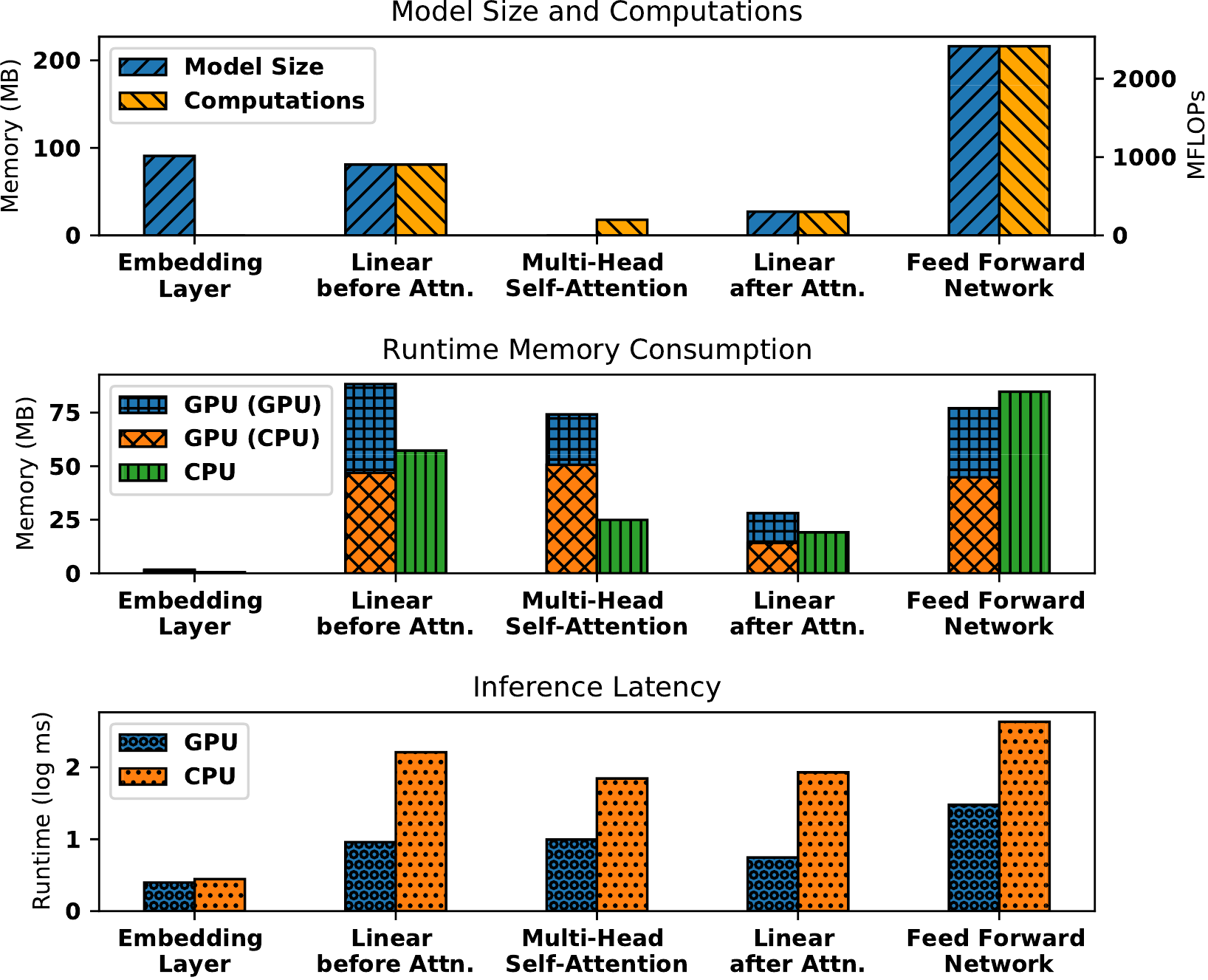}
\caption{Breakdown Analysis of $\mathrm{BERT_{BASE}}$.}
\label{fig:memory}
\end{figure*}

As shown in Fig.~\ref{fig:bert_breakdown}, the Transformer backbone has multiple stacked encoder units, each with two major sub-units: a self-attention sub-unit and a feed forward network (FFN) sub-unit, both with residual connections. 
Each self-attention sub-unit consists of a multi-head self-attention layer, and fully-connected layers before and after it.
An FFN sub-unit exclusively contains fully-connected layers.
The architecture of BERT can be specified using three hyper-parameters: number of encoder units ($L$), size of the embedding vector ($H$), and number of attention heads in each self-attention layer ($A$). $L$ and $H$ determine the depth and the width of the model, whereas $A$ is an internal hyper-parameter that affects the number of contextual relations that each encoder can focus on.
The authors of BERT provide two pre-trained models: $\mathrm{BERT_{BASE}}$ ($L=12;H=768;A=12$) and $\mathrm{BERT_{LARGE}}$ ($L=24;H=1024;A=16$).

We conducted various experiments on $\mathrm{BERT_{BASE}}$ model by running inference on a sentence of length 256 and then collected the results in Fig.~\ref{fig:memory}. The top graph in the figure compares the model size as well as the theoretical computational requirements (measured in millions of FLOPs) of different parts of the model.
The bottom two graphs track the model's run-time memory consumption as well as the inference latency on two representative hardware setups.
We conducted experiments using Nvidia Titan X GPU with 12GB of video RAM and Intel Xeon E5-1620 CPU with 32 GB system memory, which is a commonly used server or workstation configuration.
All data was collected using the PyTorch profiling tool.

Clearly, the parts consuming the most memory in terms of model size and executing the highest number of FLOPs are the FFN sub-units. The embedding layer is also a substantial part of the model size, due to the large vector size ($H$) used to represent each embedding vector. Note that the embedding layer has zero FLOPs, since it is a lookup table that involves no arithmetic computations at inference time. For the self-attention sub-units, we further break down the costs into multi-head self-attention layers and the linear (\textit{i.e.},~fully-connected) layers before and after them. The multi-head self-attention does not have any learnable parameters; however, its computational cost is non-zero due to the dot products and the softmax operations. 

The linear layers surrounding each attention layer incur additional memory and computational overhead, though it is relatively small compared to the FFN sub-units. Note that the input to the attention layer is divided among various heads, and thus each head operates in a lower-dimensional space ($H/A$). The linear layer before attention is roughly three times the size of that after it, since each attention has three inputs (key, value, and query) and only one output.

The theoretical computational overhead may differ from the actual inference cost at run-time, which depends on the hardware that the model runs on. 
As expected, when running the model on a GPU, the total run-time memory includes both memory on the GPU side and the CPU side, and it is greater than for a model running solely on CPU due to duplicate tensors present on both devices for faster processing on GPU. 

The most notable difference between the theoretical analysis and the run-time measurements on a GPU is that the multi-head self-attention layers are significantly more costly in practice than in theory. This is because the operations in these layers are rather complex, and are implemented as several matrix transformations followed by a matrix multiplication and softmax.
Furthermore, GPUs are designed to accelerate certain operations, and can thus implement linear layers faster and more efficiently than the more complex attention layers.
When we compare the run-time performance on a CPU, where the hardware is not specialized for linear layer operations, the inference time as well as the memory consumption of all the linear layers shoots up more compared to the multi-head self-attention. Thus on a CPU, the behavior of run-time performance is similar to that of theoretical computations. The total execution time of a single example on a GPU (57.1 ms) is far superior as compared to a CPU (750.9 ms), as expected.
The execution time of the embedding layer is largely independent of the hardware on which the model is executed (since it is just a table lookup) and it is relatively small compared to other layers.
The FFN sub-units are the bottleneck of the whole model, which is consistent with the results from the theoretical analysis.

%% file: sections/03-Compression_Methods.tex
\section{Compression Methods}\label{sec:compression}

Due to BERT's complex architecture, no existing compression method focuses on every aspect of the model like self-attention, linear layers, embedding size, model depth, \textit{etc.} Instead, each compression technique applies to certain components of BERT. Below, we consider the compression methods that provide model size reduction and speedup at inference time, rather than the training procedure.

\subsection{Quantization}
\label{DQ}

Quantization refers to reducing the number of unique values required to represent the model weights, which in turn allows to represent them using fewer bits, to reduce the memory footprint, and to lower the precision of the numerical calculations. 
Quantization may even improve the runtime memory consumption as well as the inference speed when the underlying computational device is optimized to process lower-precision numerical values, \textit{e.g.},~tensor cores in newer generations of Nvidia GPUs. Programmable hardware such as FPGAs can also be specifically optimized for any bitwidth representation. Quantization of intermediate outputs and activations can further speed up the model execution \cite{boo2020fixed}.

\begin{figure}
\centering
\includegraphics[width = 0.48\textwidth]{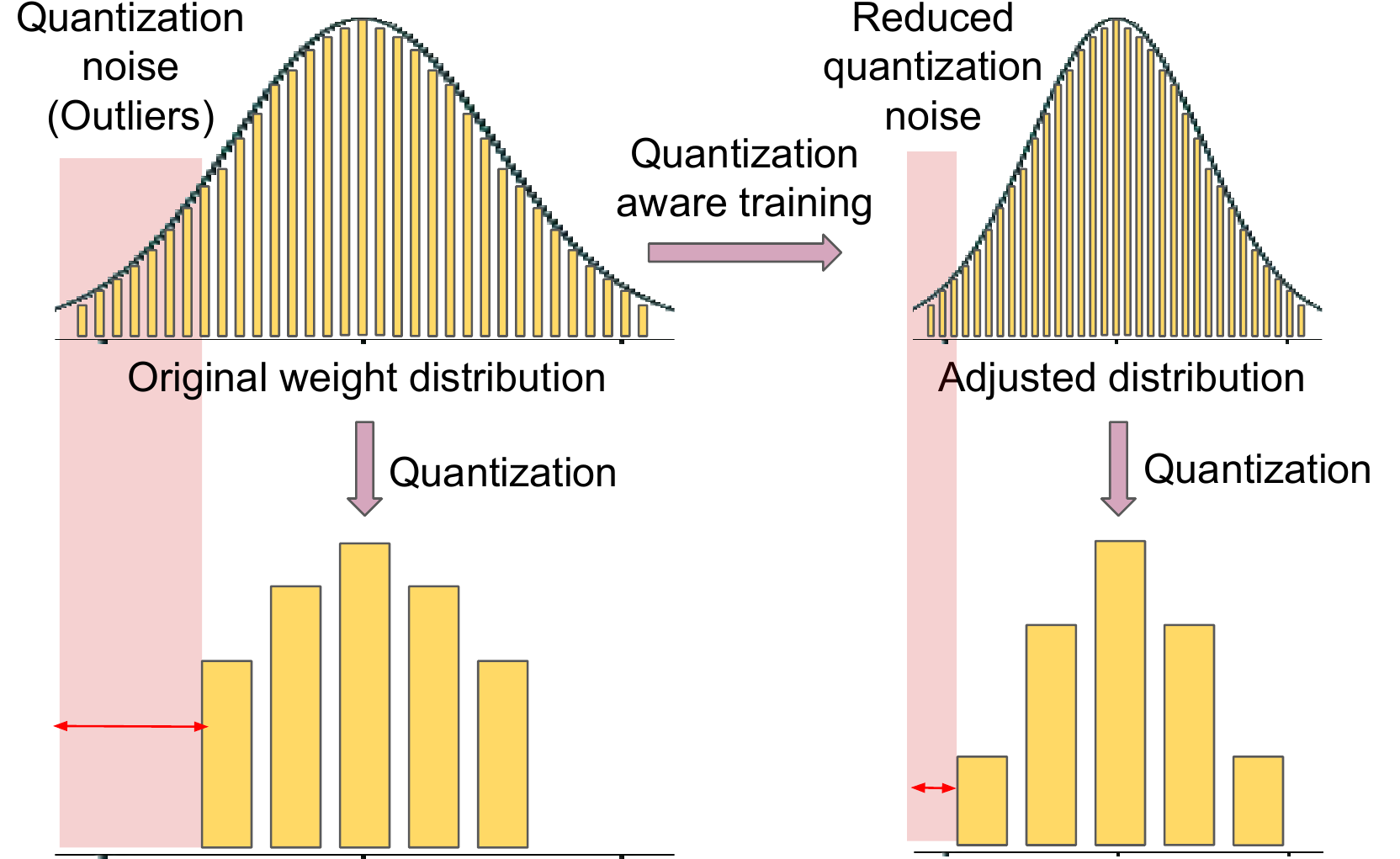}
\caption{Quantization.}
\label{fig:quant}
\end{figure}

Quantization is generally applicable to all model weights as the BERT weights reside in fully-connected layers (\textit{i.e.},~the embedding layer, the linear layers, and the FFN sub-units), which have been shown to be quantization-friendly~\cite{hubara2017quantized}. The original BERT model provided by Google represents each weight by a 32-bit floating point number. A na\"{i}ve approach is to simply truncate each weight to the target bitwidth, which often yields a sizable drop in accuracy as this forces certain weights to go through a severe drift in their value, known as quantization noise \cite{fan2020training}. A possible way around is to identify these weights and then not truncate them during the quantization step in order to retain the model accuracy. For example, \citet{zadeh2020gobo} assume Gaussian distribution in the weight matrix and identify outliers. Then, by not quantizing the outliers, they are able to perform post-training quantization without any retraining requirements.

A more common approach to retaining model accuracy is Quantization-Aware Training (QAT), which involves additional training steps to adjust the quantized weights. Fig.~\ref{fig:quant} shows an example of na\"{i}ve linear quantization, quantization noise, and the importance of quantization-aware training. For BERT, QAT has been used to perform fixed-length integer quantization~\cite{zafrir2019q8bert, boo2020fixed}, Hessian-based mixed-precision quantization~\cite{shen2019q}, adaptive floating-point quantization~\cite{tambe2020edgebert}, and noise-based quantization~\cite{fan2020training}. Finally, it has been observed that the embedding layer is more sensitive to quantization than other encoder layers, and requires more bits to maintain the model accuracy~\cite{shen2019q}.

\subsection{Pruning} \label{sec:pruning}
Pruning refers to identifying and removing redundant or less important weights and/or components, which sometimes even makes the model more robust and better-performing. Moreover, pruning is a commonly used method of exploring the lottery ticket hypothesis in neural networks \cite{frankle2018lottery}, which has also been studied in the context of BERT \cite{chen2020lottery, prasanna2020bert}. Pruning methods for BERT largely fall into two categories.

\begin{figure*}
\centering
\includegraphics[width = 0.8\textwidth]{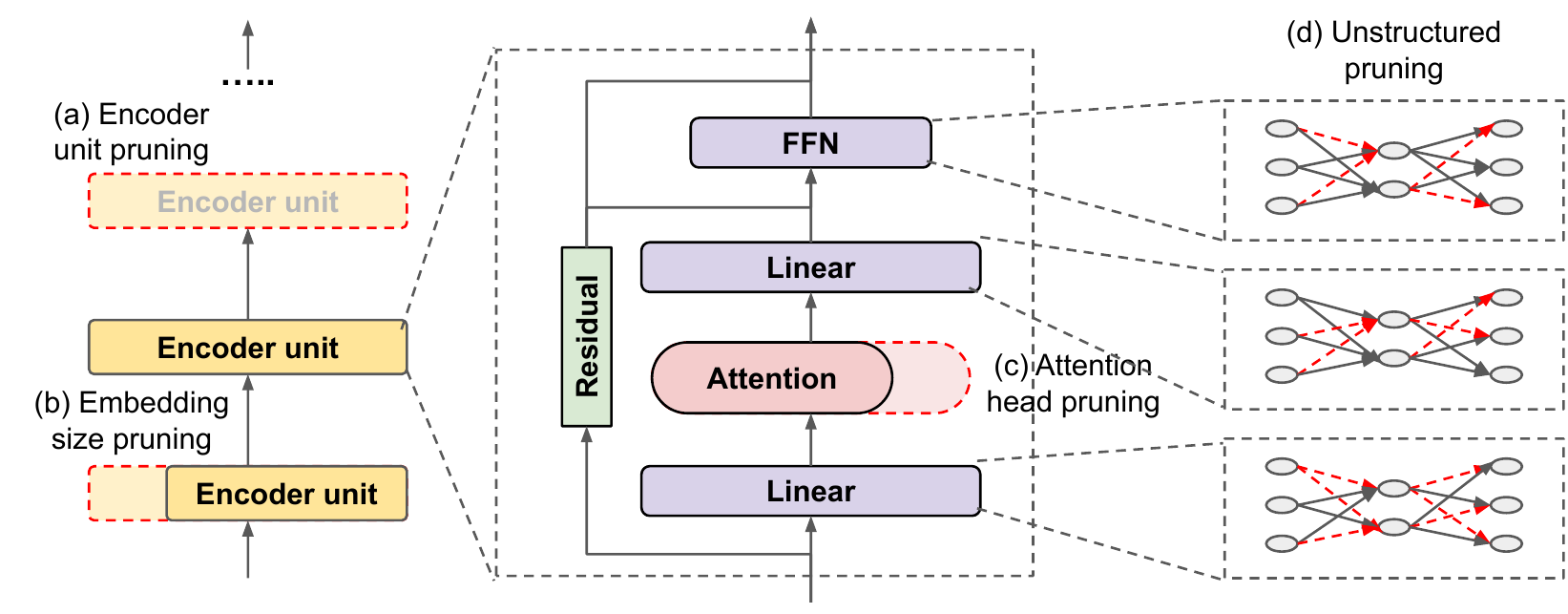}
\caption{Various pruning methods including structured pruning by (a) pruning encoder units (L), (b) pruning embedding size (H), (c) pruning attention heads (A), as well as (d) unstructured pruning.}
\label{fig:pruning}
\end{figure*}

\textbf{Unstructured Pruning.}
Unstructured pruning, also known as sparse pruning, prunes individual weights by locating the set of least important weights in the model. The importance of weights can be judged by their absolute values, gradients, or a custom-designed measurement~\cite{gordon2019,mao2020ladabert,guo2019reweighted, sanh2020movement, chen2020lottery}. Unstructured pruning could be effective for BERT, given the latter's massive fully-connected layers. Unstructured pruning methods include magnitude weight pruning \cite{gordon2019,mao2020ladabert, chen2020lottery}, which simply removes weights close to zero, movement-based pruning \cite{sanh2020movement, tambe2020edgebert}, which removes weights moving towards zero during fine-tuning, and reweighted proximal pruning (RPP) \cite{guo2019reweighted}, which uses iteratively reweighted ${\ell}_{1}$ minimization followed by the proximal algorithm for decoupling pruning and error back-propagation. Since unstructured pruning considers each weight individually, the set of pruned weights can be arbitrary and irregular, which in turn might decrease the model size, but with negligible improvement in runtime memory or speed, unless executed on specialized hardware or with specialized processing libraries.

\textbf{Structured Pruning.} Unlike unstructured pruning, structured pruning focuses on pruning structured blocks of weights \cite{li2020efficient} or even complete architectural components in the BERT model, by reducing and simplifying certain numerical modules:
\begin{itemize}[noitemsep, leftmargin=*]
    \item \emph{Attention head pruning.} The self-attention layer incurs considerable computational overhead at inference time; yet, its importance has often been questioned~\cite{kovaleva2019revealing, tay2020synthesizer, aless2020fixed}. In fact, high accuracy is possible with only 1--2 attention heads per encoder unit, even though the original model has 16 attention heads~\cite{michel2019sixteen}. Randomly pruning attention heads during the training phase has been proposed, which can create a model robust to various numbers of attention heads, and a smaller model can be directly extracted for inference based on deployment requirements \cite{hou2020dynabert}.
    
    \item \emph{Encoder unit pruning.} Another structured pruning method aims to reduce the number of encoder units $L$ by pruning the less important layers. For instance, layer dropout drops encoder units randomly or with a pre-defined strategy during training. If the layers are dropped randomly, a smaller model of any desired depth can be extracted during inference~\cite{fan2019reducing,hou2020dynabert}. Otherwise, a smaller model of fixed depth is obtained \cite{sajjad2020poor,xu2020bert}. As BERT contains residual connections for every sub-unit, using an identity prior to prune these layers has also been proposed \cite{lin2020pruning}.
    
    \item  \emph{Embedding size pruning.} Similar to encoder unit pruning, we can reduce the size of the embedding vector ($H$) by pruning along the width of the model. Such a model can be obtained by either training with adaptive width, so that the model is robust to such pruning during inference ~\cite{hou2020dynabert}, or by removing the least important feature dimensions iteratively \cite{khetan2020schubert, prasanna2020bert, tsai2020finding,lin2020pruning}.
\end{itemize}

Fig.~\ref{fig:pruning} shows a visualisation of various forms of structured pruning and unstructured pruning.

\subsection{Knowledge Distillation}
\label{KD}
Knowledge Distillation refers to training a smaller model (called the \emph{student}) using outputs (from various intermediate functional components) of one or more larger pre-trained models (called the \emph{teachers}).  The flow of information can sometimes be through an intermediate model (commonly known as \emph{teaching assistants}) \cite{ding2020sdsk2bert,sunmobilebert,wang2020minilm}. In the BERT model, there are multiple intermediate results that the student can learn from, such as the logits in the final layer, the outputs of the encoder units, and the attention maps.  Moreover, there are multiple forms of loss functions adapted such as cross-entropy loss, KL divergence, MAE, etc. While knowledge distillation is most commonly used to train student models directly on task-specific data, recent results have shown that distillation during both pre-training and fine-tuning can help create better performing models \cite{song2020lightpaff}. An overview of various forms of knowledge distillation and student models is shown in Fig.~\ref{fig:distil}. Based on what the student learns from the teacher, we categorize the existing methods as follows:

\begin{figure*}
\centering
\includegraphics[width = 0.98\textwidth]{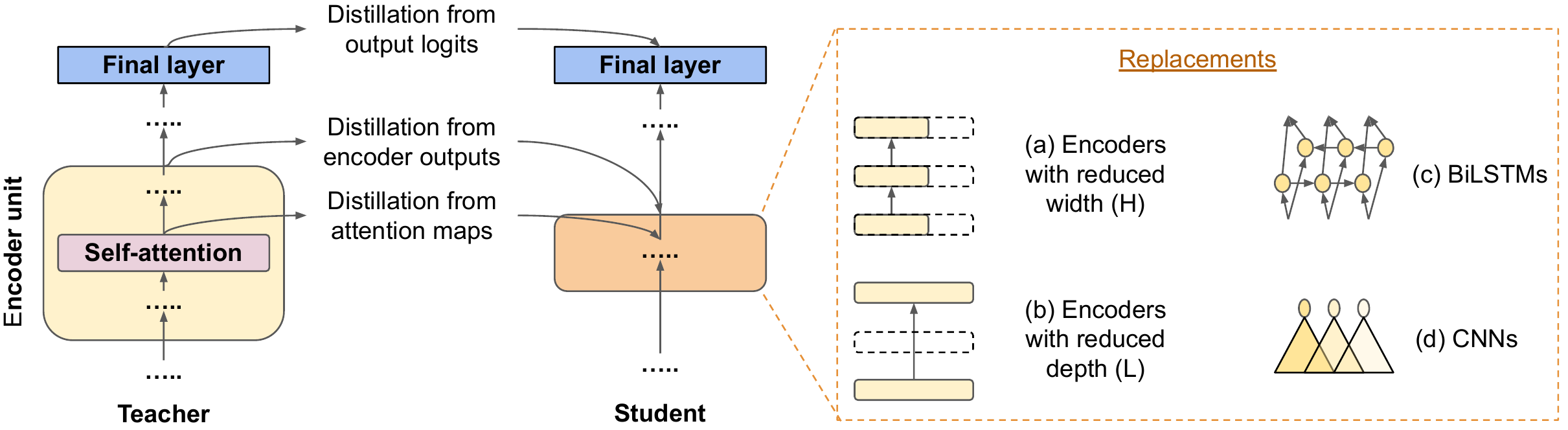}
\caption{Knowledge Distillation. Student models can be formed by (a) reducing the encoder width, (b) reducing the number of encoder, (c) replacing with a BiLSTM, (d) replacing with a CNN, or some combination of all.}
\label{fig:distil}
\end{figure*}

\textbf{Distillation from Output Logits.}
\label{subsec:ol}
Similar to knowledge distillation for CNNs~\cite{cheng2017survey}, the student can directly learn from the output logits ({\it i.e.},~from soft labels) of the final softmax layer in BERT. This is done to allow the student to better mimic the output of the teacher model, by replicating the probability distribution across various classes. 

While knowledge distillation on output logits is most commonly used to train smaller BERT models \cite{sun2019patient,sanh2019distilbert, jiao2019tinybert,zhao2019extreme,turc2019well,cao2019,sunmobilebert,song2020lightpaff,mao2020ladabert,wu2020distilling,li2020bert,ding2020sdsk2bert,noach2020compressing}, the student does not need to be a smaller version of BERT or even a Transformer, and can follow a completely different architecture. Below we describe the two commonly used replacements:

\begin{itemize}[noitemsep, leftmargin=*]
    \item \emph{Replacing the Transformer with a BiLSTM}, to create a lighter backbone. Recurrent models such as BiLSTMs process words sequentially instead of simultaneously attending to each word in the sentence like Transformers do, resulting in a smaller runtime memory requirement. Both can create bidirectional representations, and thus BiLSTMs can be considered a faster alternative to Transformers \cite{wasserblat2020exploring}. Compressing to a BiLSTM is typically done directly for a specific NLP task \cite{mukherjee2020xtremedistil}. Since these models are trained from scratch on the task-specific dataset without any intermediate guidance, various methods have been proposed to create additional synthetic training data using rule-based data augmentation techniques \cite{tang2019distilling,tang2019natural, mukherjee2019distilling} or to collect data from multiple tasks to train a single model \cite{liu2019attentive}.
    
    \item \emph{Replacing the Transformer with a CNN}, to take advantage of massively parallel computations and improved inference speed~\cite{chia2019transformer}. While it is theoretically possible to make the internal processing of an encoder parallel, where each parallel unit requires access to all the inputs from the previous layer as an encoder unit focuses on the global context, this setup is computationally intensive and cost-inefficient. Unlike Transformers, each CNN unit focuses on the local context only, and, unlike BiLSTMs, CNNs do not operate on the input sequentially, which makes it easier for them to divide the computation into small parallel units. It is possible to either completely replace the Transformer backbone with a deep CNN network \cite{chen2020adabert}, or to replace only a few encoder units to balance performance and efficiency \cite{tian2019waldorf}.

\end{itemize}

\textbf{Distillation from Encoder Outputs.}  Each encoder unit in a Transformer model can be viewed as a separate functional unit. Intuitively, the output tensors of such an encoder unit may contain meaningful semantic and contextual relationships between input tokens, leading to an improved representation. Following this idea, we can create a smaller model by learning from an encoder's outputs. The smaller model can have a reduced embedding size $H$, a smaller number of encoder units $L$, or a lighter alternative that replaces the Transformer backbone.

\begin{itemize}[noitemsep, leftmargin=*]
    \item Reducing $H$ leads to more compact representations in the student  \cite{zhao2019extreme,sunmobilebert,jiao2019tinybert,li2020bert}. One challenge is that the student cannot directly learn from the teacher's intermediate outputs, due to different sizes. To overcome this, the student also learns a transformation, which can be implemented by either down-projecting the teacher's outputs to a lower dimension or by up-projecting the student's outputs to the original dimension \cite{zhao2019extreme}.  Another possibility is to introduce these transformations directly into the student model, and later merge them with the existing linear layers to obtain the final smaller model \cite{zhou2020go}.

    \item Reducing $L$, which is the number of encoder units, forces each encoder unit in the student to learn from the behavior of a sequence of multiple encoder units in the teacher~\cite{sun2019patient,sanh2019distilbert,sunmobilebert,jiao2019tinybert,zhao2019extreme,li2020bert}. Further analysis into various details of choosing which encoder units to use for distillation is provided by \citet{sajjad2020poor}. For example, preserving the bottom encoder units and aggressively distilling the top encoder units yields a better-performing student model, which indicates the importance of the bottom layers in the teacher model. 
     While most existing methods create an injective mapping from the student encoder units to the teacher, \citet{li2020bert} instead propose a way to build a many-to-many mapping for a better flow of information. One can also completely bypass the mapping by combining all outputs into one single representation vector \cite{sun2020contrastive}.
    
    \item It is also possible to use encoder outputs to train student models that are not Transformers \cite{mukherjee2019distilling, mukherjee2020xtremedistil, tian2019waldorf}. However, when the student model uses a completely different architecture, the flexibility of using internal representations is rather limited, and only the output from the last encoder unit is used for distillation.
\end{itemize}

\textbf{Distillation from Attention Maps.}
Attention map refers to the softmax distribution output of the self-attention layers and indicates the contextual dependence between the input tokens. It has been proposed that attention maps in BERT can identify distinguishable linguistic relations, \textit{e.g.}, identical words across sentences, verbs and corresponding objects, or pronouns and corresponding nouns~\cite{clark2019does}. These distributions are the only source of inter-dependency between input tokens in a Transformer model and thus by replicating these distributions, a student can also learn such linguistic relations~\cite{sunmobilebert,jiao2019tinybert,mao2020ladabert,tian2019waldorf,li2020bert,noach2020compressing}. 

A common method of distillation from attention maps is to directly minimize the difference between the teacher and the student multi-head self-attention outputs. Similar to distillation from encoder outputs, replicating attention maps also faces a choice of mapping between the teacher and the student, as each encoder unit has its own attention distribution. Previous work has also proposed replicating only the last attention map in the model to truly capture the contextual dependence \cite{wang2020minilm}.
One can attempt an even deeper distillation of information through intermediate attention outputs such as key, query and value matrices, individual attention head outputs, key-query and value-value matrix products, \textit{etc.}, to facilitate the flow of information ~\cite{wang2020minilm,noach2020compressing}.

\subsection{Matrix Decomposition}

The computational overhead in BERT mainly consists of large matrix multiplications, both in the linear layers as well as in the attention heads. Thus, decomposing these matrices can significantly impact the computational requirement for such models.

\textbf{Weight Matrix Decomposition.} 
A possible way to reduce the computational overhead of the model can be through weight matrix factorization, which replaces the original $A \times B$ weight matrix with the product of two smaller matrices ($A \times C$ and $C \times B$). The reduction in model size as well as runtime memory usage is sizable if $C \ll A,B$. The method can be used to reduce the model size and the computations for linear layers in the model \cite{noach2020compressing,mao2020ladabert}, as well as the embedding matrix \cite{lan2019albert,tambe2020edgebert}.

\begin{figure}
\centering
\includegraphics[width = 0.48\textwidth]{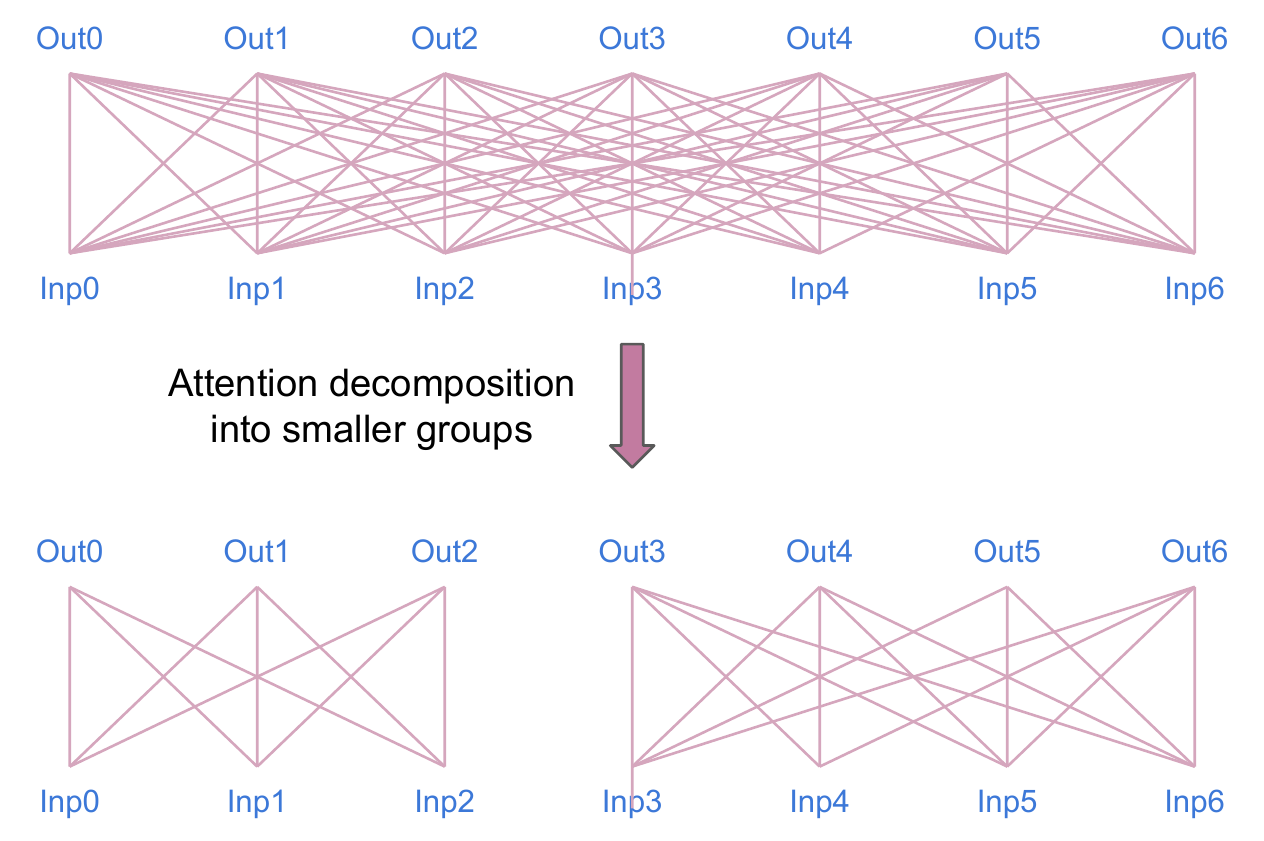}
\caption{Attention decomposition.}
\label{fig:attn_decom}
\end{figure}

\textbf{Attention Decomposition.}
The importance of attention calculation over the entire sentence has been explored, revealing a large number of redundant computations \cite{tay2020synthesizer, cao2019}. One way to resolve this issue is by calculating attention in smaller groups, by either binning them using spatial locality \cite{cao2019}, magnitude-based locality \cite{kitaev2019reformer}, or an adaptive attention span \cite{tambe2020edgebert}. Moreover, since the outputs are calculated independently, local attention methods also enable a higher degree of parallel processing and individual representations can be saved during inference for multiple uses. Fig.~\ref{fig:attn_decom} shows an example of attention decomposition based on spatial locality.

It has been also proposed to reduce the computations required in the attention calculation by projecting the key-query matrix into a lower dimensionality \cite{wang2020linformer} or by only calculating the softmax of the top-$k$ key-query product values in order to further highlight these relations \cite{zhao2019explicit}. Since the multi-head self-attention layer does not contain weights, these methods only improve the runtime memory costs and execution speed, but not the model size.

\subsection{Dynamic Inference Acceleration}

\begin{figure}
\centering
\includegraphics[width = 0.45\textwidth]{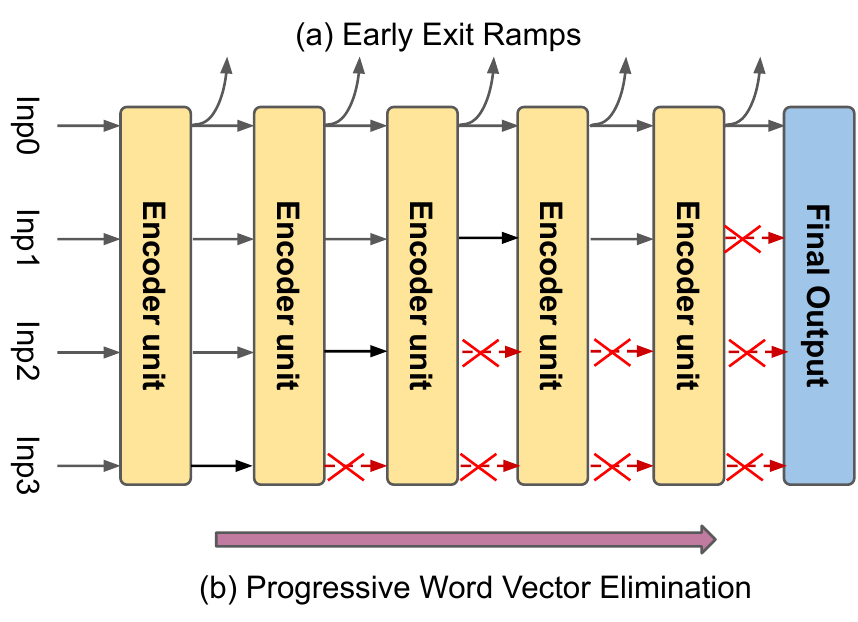}
\caption{Dynamic inference acceleration.}
\label{fig:dynamic}
\end{figure}

Besides directly compressing the model, there are methods that focus on reducing computational overhead at inference time, catering to individual input examples and dynamically changing the amount of computation required. We have visualized various dynamic inference acceleration methods in Fig.~\ref{fig:dynamic}.

\textbf{Early Exit Ramps.}
One way to speed up inference is to create intermediary exit points in the model. Since the classification layers are the least parameter-extensive part of BERT, separate classifiers can be trained for each encoder unit output. This allows the model to get dynamic inference time for various inputs. Training these classifiers can either be done from scratch \cite{xin2020deebert,zhou2020bert,tambe2020edgebert} or through distilling the output from the final classifier \cite{liu2020fastbert}.

\textbf{Progressive Word Vector Elimination.}
Another way to accelerate inference is by reducing the number of words processed at each encoder level. Since we use only the final output corresponding to the [CLS] token (defined in Section \ref{sec:breakdown}) as a representation of the complete sentence, the information of the entire sentence must have fused into that one token. \citet{goyalpower} observe that such a fusion cannot be sudden, and that it must happen progressively across various encoder levels. We can use this information to lighten the later encoder units by reducing the sentence length through word vector elimination at each step.

\subsection{Other Methods}

Besides the aforementioned categories, there are also several one-of-a-kind methods that have been shown to be effective for reducing the size and inference time of BERT-like models.

\textbf{Parameter Sharing.}
ALBERT \cite{lan2019albert} follows the same architecture as BERT, with the important difference that in the former, weights are shared across all encoder units, which yields a significantly reduced memory footprint. Moreover, ALBERT has been shown to enable training larger and deeper models. For example, while BERT's performance peaks at $\mathrm{BERT_{LARGE}}$ (performance of $\mathrm{BERT_{XLARGE}}$ drops significantly), the performance of ALBERT keeps improving until the far larger $\mathrm{ALBERT_{XXLARGE}}$ model ($L=12;H=4096;A=64$).

\textbf{Embedding Matrix Compression.} 
The embedding matrix is the lookup table for the embedding layer, which is about 21\% of the size of the complete BERT model. One way to compress the embedding matrix is by reducing the vocabulary size $V$, which is about 30k in the original BERT model. Recall from Section \ref{sec:breakdown} that the vocabulary of BERT is learned using a WordPiece tokenizer. 
The WordPiece tokenizer relies on the vocabulary size to figure out the degree of fragmentation of words present in the input text. A large vocabulary size allows for better representation of rare words and for more adaptability to OOV words. 

However, even with a 5k vocabulary size, 94\% of the tokens created match with the tokens created using a 30k vocabulary size \cite{zhao2019extreme}. This shows that the majority of words that appear frequently enough in text are covered even with such a small vocabulary size, and thus makes it reasonable to consider decreasing vocabulary size to compress the embedding matrix.

Another alternative is to replace the existing one-hot vector encoding with a ``codebook''-based encoding, where each token is represented using multiple indices from the codebook. The final embedding of the token can be calculated by doing the sum of embeddings present at all these indices \cite{prakash2020compressing}.

\textbf{Weight Squeezing.}
Weight Squeezing \cite{chumachenko2020weight} is a compression method similar to knowledge distillation, where the student learns from the teacher. However, instead of learning from intermediate outputs like in knowledge distillation, it's the weights of the teacher model that are mapped to the student through a learnable transformation, and thus the student learns its weights directly from the teacher.

%% file: sections/04-Evaluation.tex
\section{Effectiveness of the Compression Methods}

In this section, we compare the performance of several BERT compression techniques based on their model size and speedup, as well as their accuracy/{F}1 on various NLP tasks. We chose papers whose results are either on the Pareto frontier \cite{deb2014multi} or representative for each compression technique mentioned in the previous section.

\subsection{Datasets and Evaluation Measures}

From the General Language Understanding Evaluation (GLUE) benchmark \cite{wang2018glue} and the Stanford Question Answering Dataset (SQuAD) \cite{rajpurkar2016squad}, we use the following most commonly reported tasks: MNLI and QQP for sentence pair classification, SST-2 for single sentence classification, and SQuAD v1.1 for machine reading comprehension. Following the official leaderboards, we report the accuracy for MNLI, SST-2 and QQP, and F1 score for SQuAD v1.1. We also report the absolute drop in accuracy with respect to $\mathrm{BERT_{BASE}}$, averaged across all tasks for which the authors have reported results, in an attempt to quantify their accuracy on a single scale.

Furthermore, we report model speedup on both GPU and CPU devices collected directly from the original papers. For papers that provide their speedup, we also mention the target device on which the speedups are calculated. For papers which do not provide speedup values, we run their models on our own machine and perform inference on the complete MNLI-m test dataset (using a batch size of 1) with machine configurations as detailed in Section~\ref{sec:breakdown}. We also provide the model size with and without the embedding matrix, since for certain application scenarios, where the memory constraints for model storage are not strict, the parameters of the embedding matrix can be ignored as it has negligible run-time cost (see Section 2).
As no previous work in the literature reports a drop in runtime memory, and as many of the papers that we compare to here use probabilistic models that cannot be easily replicated without the authors releasing their code, we could not do direct runtime memory comparisons.

\subsection{Comparison and Analysis}

\input{tables/table_eval_new.tex}

Table \ref{tab:evaluation_small} compares the effectiveness of various BERT compression methods. Note that some methods compress only part of the model, however, for uniformity, all model sizes and speedups reported here are for the final complete model after compression. Thus, certain values might not match exactly what is reported in the original papers. Below, we describe several interesting trends in Table \ref{tab:evaluation_small}.

\textbf{Quantization and Pruning.} Quantization is well-suited for BERT, and it is able to outperform other compression methods in terms of both model size and accuracy. As shown in Table~\ref{tab:evaluation_small}, quantization can reduce the size of the BERT model to 15\% and 10.2\% of its original size, with accuracy drop of only 0.6\% and 0.9\%, respectively, across various tasks \cite{shen2019q,zadeh2020gobo}. This can be attributed to the fact that quantization is an architecture-invariant compression method, which means it only reduces the precision of the weights, but preserves all the original components and connections present in the model.
Unstructured pruning also shows performance that is on par with other methods. We can see that unstructured pruning reduces the original BERT model to 67.6\% of its original size, without any loss in accuracy, possibly due to the regularization effect caused by pruning \cite{guo2019reweighted}.
However, almost all existing work in unstructured pruning freezes the embedding matrix and focuses only on pruning weight matrices of the encoder. This makes extreme compression difficult, e.g., even with 3\% weight density in encoders, the total model size still remains at 23.8\% of its original size \cite{sanh2020movement}, and yields a sizable drop in accuracy/F1 (4.73\% on average).

While both quantization and unstructured pruning reduce the model size significantly, none of them yields actual run-time speedups on a standard device. Instead, specialized hardware and/or libraries are required, which can do lower-bit arithmetic for quantization and an optimized implementation of sparse weight matrix multiplication for unstructured pruning. However, these methods can be easily combined with other compression methods as they are orthogonal from an implementation viewpoint. We discuss the performance of compounding multiple compression method together later in this section.

\textbf{Structured Pruning.} As discussed in Section \ref{sec:compression}, structured pruning removes architectural components from BERT, which can also be seen as reducing the hyper-parameters that govern the BERT architecture. While \citet{lin2020pruning} pruned the encoder units ($L$) and reduced the model depth by half with an average accuracy drop of 1.0\%, \citet{khetan2020schubert} took it a step further and systematically reduced both the depth ($L$) as well as the width ($H$, $A$) of the model, and were able to compress it to 39.1\% of its original size with an average accuracy drop of only 1.86\%. Detailed experiments by \citet{khetan2020schubert} also show that reducing all hyper-parameters in harmony, instead of focusing on just one, helps achieve better performance.

\textbf{Model-Agnostic Distillation.} Applying distillation from output logits only allows model-agnostic compression and gives rise to LSTM/CNN-based student models. While some methods exist that do try to train a smaller BERT model \cite{song2020lightpaff}, this category is dominated by methods that aim to replace Transformers with lighter alternatives. It is also interesting to note that at a similar model size, \citet{liu2019attentive} which uses a BiLSTM student model, yields significantly better speedup when compared to \citet{song2020lightpaff}, which uses a Transformer-based student model. \citet{chen2020adabert} provides the fastest model in this category, using a NAS-based CNN model, with only 2.06\% average drop in accuracy.

While these methods focused on achieving high compression ratio, they cause a sizable drop in accuracy. A possible explanation is that the total model size is not a true indicator of how powerful these compression methods are, because the majority of their model size is the embedding matrix. For example, while the total model size of \citet{liu2019attentive} is 101 MB, only 11 MB is actually their BiLSTM model, and the remaining 90 MB are just the embedding matrix. Similarly to unstructured pruning, not focusing on the embedding matrix can hurt the deployment of such models on devices with strict memory constraints.

\textbf{Distillation from Attention Maps.}
\citet{wang2020minilm} were able to reduce the model to 60.7\% its original size, with only 0.1\% loss in accuracy on average, just by doing deep distillation on the attention layers. For the same student architecture, \citet{sanh2019distilbert} used all other forms of distillation (\emph{i.e.},~output logits and encoder outputs) together and still faced an average accuracy loss of 1.73\%. Clearly, the intermediate attention maps are an important distillation target.

\textbf{Combining Multiple Distillations.}
Combining multiple distillation targets can help achieve an even better performing compressed model. \citet{jiao2019tinybert} created a student model with smaller $H$ and $L$ hyperparameter values, and were able to compress the model size to 13.3\% and achieved a 9.4x speedup on GPU (9.3x on CPU), while only facing a drop of 1.0\% in accuracy.
\citet{zhao2019extreme} extended the same idea and created an extremely small BERT student model (1.6\% of the original size and $\sim$ 25x faster) with $H = 48$ and vocabulary size $|V| = 4928$ ($\mathrm{BERT_{BASE}}$ has $H = 768$ and $|V| = 30522$). The model lost 12.3\% accuracy to make up for its size.

\textbf{Matrix Decomposition and Dynamic Inference Acceleration.}
While weight matrix decomposition helps reduce the size of the weight matrices in BERT, they create deeper and fragmented models, which hurts the execution time \cite{noach2020compressing}. On the other hand, methods from the literature that implement faster attention and various forms of dynamic speedup do not change the model size, but instead provide faster inference. For example, the work by \citet{cao2019} show that attention calculation across the complete sentence is not needed for the initial encoder layers, and they were able to achieve $\sim$ 3x run-time speedup with only 0.76\% drop in accuracy. For applications where latency is the major constraint, such methods can help achieve the necessary speedup.

\textbf{Structured Pruning vs. Distillation.} While structured pruning attempts to iteratively prune the hyper-parameters of BERT, distillation starts with a smaller model and tries to train it using knowledge directly from the original BERT. However, both of them end up with a similar compressed model, and thus it is interesting to compare which path yields more promising results. As can be noted from Table~\ref{tab:evaluation_small}, for the same compressed model with $L = 6$, the drop in accuracy for the model by \citet{lin2020pruning} is smaller compared to that by \citet{sanh2019distilbert}. However, this is not a completely fair comparison, as \citet{sanh2019distilbert} does not use attention as a distillation target. When we compare other methods, we find that \citet{jiao2019tinybert} was able to beat \citet{khetan2020schubert} in terms of both model size and accuracy. This shows that structured pruning outperforms student models trained using distillation only on encoder outputs and output logits, but fails against distillation on attention maps. This further indicates the importance of replicating attention maps in BERT.

\textbf{Pruning with Distillation.} Similar to combining multiple distillation methods, one can also combine pruning with distillation, as this can help guide the pruning towards removing less important connections. \citet{mao2020ladabert} combined distillation with unstructured pruning, while \citet{hou2020dynabert} combined distillation with structured pruning. When compared with only structured pruning \cite{khetan2020schubert}, we see that \citet{hou2020dynabert} achieved a smaller model size (12.4\%), and a smaller accuracy drop (0.96\%).

\textbf{Quantization with Distillation.} Similar to pruning, quantization is also orthogonal in implementation to distillation, and can together achieve better performance than either of them individually. \citet{zadeh2020gobo} attempted to quantize an already distilled BERT \cite{sanh2019distilbert} to 4 bits, thus reducing the model size from 60.2\% to 7.5\%, with an additional accuracy drop of only 0.9\% (1.73\% to 2.6\%). Similarly, \citet{sunmobilebert} attempted to quantize their model to 8 bits, which reduced their model size from 23\% to 5.25\%, with only a 0.07\% additional drop in accuracy.

\textbf{Compounding multiple methods together.}
As we have seen in this section, different methods of compression target different parts of the BERT architecture. Note that many of these methods are orthogonal in implementation, similar to the papers that we discussed on combining quantization and pruning with distillation, and thus it is possible to combine them. For example, \citet{tambe2020edgebert} combined multiple forms of compression methods to create a truly deployable language model for edge devices. They combined parameter sharing, embedding matrix decomposition, unstructured movement pruning, adaptive floating-point quantization, adaptive attention span, dynamic inference speed with early exit ramps, and other hardware accelerations to suit their needs. However, as we noticed in this section, these particular methods can reduce the model size significantly, but they cannot drastically speed up the model execution on standard devices. While the model size is dropped to only 1.3\% of its original size, the speedup obtained on a standard GPU is only 1.83x, with an average drop of 1.53\% in accuracy. With specialized accelerators, the authors were able to push the speedup eventually to 2.1x.

\subsection{Practical Advice}

Based on the results collected in this section, we attempt to give the readers some practical advice for specific applications:

\begin{itemize}[noitemsep, leftmargin=*]
    \item Quantization and unstructured pruning can help reduce the model size, but do nothing to improve the runtime inference speed or the memory consumption, unless executed on specialized hardware or with specialized processing libraries. On the other hand, if executed on proper hardware, these methods can provide tremendous boost in terms of speed with negligible loss in performance \cite{zadeh2020gobo,tambe2020edgebert,guo2019reweighted}. Thus, it is important to recognize the target hardware device before using these compression methods.
    
    \item Knowledge distillation has shown great affinity to a variety of student models and its orthogonal nature of implementation compared to other methods \cite{mao2020ladabert,hou2020dynabert} means it is an important addition to any form of compression. More specifically, distillation from self-attention layers (if possible) is an integral part of Transformer compression \cite{wang2020minilm}.
    
    \item Alternatives such as BiLSTMs and CNNs have an additional advantage in terms of execution speed when compared to Transformers. Thus, for applications with strict latency constraints, replacing Transformers with alternative units is a better choice. Model execution can also be sped up using dynamic inference methods, as they can be incorporated into any student model with a skeleton similar to that of Transformers.
    
    \item A major takeaway is the importance of compounding various compression methods together to achieve truly practical models for edge environment. The work of \citet{tambe2020edgebert} is a good example of this, as it attempts to compress BERT, while simultaneously performing hardware optimizations in accordance with their chosen compression methods. Thus, combining compression methods that complement each other is better than compressing a single aspect of the model to its extreme.
\end{itemize}

%% file: tables/table_eval_new.tex
\begin{table*}[!h]
\footnotesize
\centering
\setlength\tabcolsep{1.5pt}
\begin{tabular}{|c|c|c|c|c|c|c|c|c|c|c|c|}
\Xhline{3\arrayrulewidth}
 \textbf{Methods} & \textbf{Provenance} & \textbf{Target} & \multicolumn{2}{|c|}{\textbf{Model Size}} & \multicolumn{2}{|c|}{\textbf{Speedup}} & \multicolumn{4}{|c|}{\textbf{Accuracy/F1}} & \textbf{Avr.} \\
\cline{4-11}
 & & \textbf{Device} & \textbf{w/ emb} & \textbf{w/o emb} & \textbf{GPU} & \textbf{CPU} & \textbf{MNLI} & \textbf{QQP} & \textbf{SST-2} & \textbf{SQD} & \textbf{Drop} \\
\Xhline{3\arrayrulewidth}

$\mathrm{BERT}_{\mathrm{BASE}}$ & \cite{devlin2019bert} & -- & 100\% & 100\% & 1x & 1x & 84.6 & 89.2 & 93.5 & 88.5 & 0.0 \\
\Xhline{3\arrayrulewidth}

\multirow{2}{*}{Quantization} & \cite{shen2019q} $^{S}$ & -- & 15\% & 12.5\% & 1x & 1x & 83.9 & -- & 92.6 & 88.3 & -0.6 \\
\cline{2-12}
 & \cite{zadeh2020gobo} $^{S}$ & -- & 10.2\% & 5.5\% & 1x & 1x & 83.7 & -- & -- & -- & -0.9 \\
% \Xhline{3\arrayrulewidth}

\Xhline{3\arrayrulewidth}

\multirow{2}{*}{Unstructured} & \cite{guo2019reweighted} $^{A}$ & -- & 67.6\% & 58.7\% & 1x & 1x & -- & -- & -- & 88.5 & 0.0 \\
\cline{2-12}
\multirow{2}{*}{Pruning} & \cite{chen2020lottery} $^{S}$ & -- & 48.9\%$^{*}$ & 35.1\%$^{*}$ & 1x & 1x & 83.1 & 89.5 & 92.9 & 87.8 & -0.63 \\
\cline{2-12}
 & \cite{sanh2020movement} $^{S}$ & -- & 23.8\% & 3\% & 1x & 1x & 79.0 & 89.3 & -- & 79.9 & -4.73 \\

\Xhline{3\arrayrulewidth}
Structured & \cite{lin2020pruning} $^{S}$ & -- & 60.7\% & 50\% & -- & -- & -- & 88.9 & 91.8 & -- & -1.0 \\
\cline{2-12}
Pruning & \cite{khetan2020schubert} $^{A}$ & -- & 39.1\% & 38.8\% & 2.93x$^{\ddagger}$ & 2.76x$^{\ddagger}$ & 83.4 & -- & 90.9 & 86.7 & -1.86 \\
% \cline{2-12}
% & \cite{sun2020contrastive} $^{A,S}$ $^{\dagger\dagger}$ $^{\dagger}$ & 255 (60.7\%) & 165 (50\%) & 1.94x & 1.73x & 83.5 & 89.1 & 93.6 & -- & -0.37 \\
% & \cite{lin2020pruning} & 45.0\% & 189 & 99 & -- & -- & -- & 87.8 & 88.4 & -- & -3.25 \\
% & \cite{fan2019reducing} $^{A,S}$ & 255 (60.7\%) & 165 (50\%) & 1.94x & 1.73x & 82.9 & -- & 92.5 & -- & -1.35 \\
\Xhline{3\arrayrulewidth}

% \multicolumn{3}{l}{Knowledge Distillation}  \\
% \Xhline{3\arrayrulewidth}
\multirow{2}{*}{KD from} & \cite{song2020lightpaff} $^{A,S}$ & V100 & 22.8\% & 10.9\% & 6.25x & 7.09x & -- & 88.6 & 92.9 & -- & -0.6 \\
\cline{2-12}
\multirow{2}{*}{Output Logits} & \cite{liu2019attentive}$^{\dagger}$ $^{S}$ & V100 & 24.1\% & 3.3\% & 10.7x & 8.6x$^{\ddagger}$ & 78.6 & 88.6 & 91.0 & -- & -3.03 \\
\cline{2-12}
 & \cite{chen2020adabert} $^{A,S}$ & V100 & 7.4\% & 4.8\% & 19.5x$^{*}$ & -- & 81.6 & 88.7 & 91.8 & -- & -2.06 \\

\Xhline{3\arrayrulewidth}

% \multirow{2}{*}{KD-EO} & 60.5\% & 2x & 82.2 & 88.5 & 91.3 & 86.9 & 16.45 \\
% \cline{2-8}
%  & 1.6\% & -- & 71.0 & -- & 82.8 & -- & 7.23 \\
% \Xhline{3\arrayrulewidth}

KD from Attn. & \cite{wang2020minilm} $^{A}$ & P100 & 60.7\% & 50\% & 1.94x & 1.73x & 84.0 & 91.0 & 92.0 & -- & -0.1 \\

\Xhline{3\arrayrulewidth}

\multirow{3}{*}{Multiple KD} & \cite{sanh2019distilbert} $^{A}$ & CPU & 60.7\% & 50\% & 1.94x & 1.73x & 82.2 & 88.5 & 91.3 & 86.9 & -1.73 \\
\cline{2-12}
\multirow{3}{*}{combined} & \cite{sunmobilebert}$^{\dagger}$ $^{A}$ & Pixel & 23.1\% & 24.8\% & 3.9x$^{\ddagger}$ & 4.7x$^{\ddagger}$ & 83.3 & -- & 92.8 & 90.0 & -0.16 \\
% mobilebert - 5.52x on pixel phone
\cline{2-12}
 & \cite{jiao2019tinybert} $^{A,S}$ & K80 & 13.3\% & 6.4\% & 9.4x & 9.3x$^{\ddagger}$ & 82.5 & 89.2 & 92.6 & -- & -1.0\\
\cline{2-12}
 & \cite{zhao2019extreme} $^{A}$ & -- & 1.6\% & 1.8\% & 25.5x$^{\ddagger}$ & 22.7x$^{\ddagger}$ & 71.3 & -- & 82.2 & -- & -12.3 \\

\Xhline{3\arrayrulewidth}

Matrix & \cite{noach2020compressing} $^{S}$ & Titan V & 60.6\% & 49.1\% & 0.92x & 1.05x & 84.8 & 89.7 & 92.4 & -- & -0.13 \\
\cline{2-12}
Decomposition & \cite{cao2019} $^{S}$ & V100 & 100\% & 100\% & 3.14x & 3.55x & 82.6 & 90.3 & -- & 87.1 & -0.76 \\ 
% \cline{2-12}
%  & \cite{wang2020linformer} $^{\dagger}$ $^{A}$ & 420 (100\%) & 330 (100\%) & 1.6x & -- & -- & 90.5 & 93.2 & -- & +0.4 \\
\Xhline{3\arrayrulewidth}

Dynamic & \cite{xin2020deebert} $^{S}$ & P100 & 100\% & 100\% & 1.25x & 1.28x$^{\ddagger}$ & 83.9 & 89.2 & 93.4 & -- & -0.26 \\
\cline{2-12}
Inference & \cite{goyalpower} $^{S}$ & K80 & 100\% & 100\% & 2.5x & 3.1x$^{\ddagger}$ & 83.8 & -- & 92.1 & -- & -1.1 \\
% \cline{2-12}
% & \cite{zhou2020bert} $^{S}$ & 420 (100\%) & 330 (100\%) & 1.62x & -- & 83.6 & -- & 92.0 & -- & -1.25 \\
\Xhline{3\arrayrulewidth}

Param. Sharing & \cite{lan2019albert} $^{A}$ & -- & 10.7\% & 8.8\% & 1.2x$^{\ddagger}$ & 1.2x$^{\ddagger}$ & 84.3 & 89.6 & 90.3 & 89.3 & -0.58 \\

\Xhline{3\arrayrulewidth}

Pruning & \cite{mao2020ladabert} $^{S}$ & -- & 40.0\% & 37.3\% & 1x & 1x & 83.5 & 88.9 & 92.8 & -- & -0.7 \\
% \cline{2-8}
%  & \cite{sanh2020movement} & 25.9\% & 1x & 80.2 & 89.3 & 82.4 & -- \\
\cline{2-12}
with KD & \cite{hou2020dynabert} $^{S}$ & K40 & 31.2\% & 12.4\% & 5.9x$^{\ddagger}$ & 8.7x$^{\ddagger}$ & 82.0 & 90.4 & 92.0 & -- & -0.96 \\

\Xhline{3\arrayrulewidth}

Quantization & \cite{zadeh2020gobo} $^{S}$ & CPU & 7.6\% & 3.9\% & 1.94x & 1.73x & 82.0 & -- & -- & -- & -2.6 \\
% \cline{2-8}
%  & \cite{sanh2020movement} & 25.9\% & 1x & 80.2 & 89.3 & 82.4 & -- \\
\cline{2-12}
with KD & \cite{sunmobilebert}$^{\dagger}$ $^{A}$ & Pixel & 5.7\% & 6.1\% & 3.9x$^{\ddagger}$ & 4.7x$^{\ddagger}$ & 83.3 & -- & 92.6 & 90.0 & -0.23 \\
% mobilebert quant pixel --  5.52x

\Xhline{3\arrayrulewidth}

Compound  & \cite{tambe2020edgebert} $^{S}$ & TX2 & 1.3\% & 0.9\% & 1.83x & -- & 84.4 & 89.8 & 88.5 & -- & -1.53 \\

\Xhline{3\arrayrulewidth}

\end{tabular}

% \footnotemark{For methods that allow varying model size for different tasks, we used the average to represent the model size and the speedup.}

% \footnotemark{$\Delta MS$ is the drop in model size, and $\Delta A$ is the drop in accuracy for the MNLI task.}
\caption{
Evaluation of various compression methods. 
$^{*}$ indicates models using task-specific sizes or speedups; average values are reported in such cases. 
% $^{\dagger}$ marks models trained using optimizations proposed by RoBERTa \cite{liu2019roberta}. 
$^{\dagger}$ represents models that use $\mathrm{BERT_{LARGE}}$ as the teacher model. 
% $^{\dagger\dagger}$ Teacher is Roberta.
$^{\ddagger}$ represents speedup values calculated by us. Empty cells in the speedup columns are for papers which do not provide a detailed architecture of their final compressed model.
$^{A}$ represents model compressed in a task-agnostic setup, i.e. requires access to pre-training dataset.
$^{S}$ represents model compressed in task-specific setup.
V100 : Nvidia Tesla V100, P100 : Nvidia Tesla P100, K80 : Nvidia Tesla K80, Titan V : Nvidia Titan V, K40 : Nvidia Tesla K40, CPU : Intel Xeon E5, TX2 : Nvidia Jetson TX2, Pixel : Google Pixel Phone.}
\label{tab:evaluation_small}
\end{table*}

%% file: sections/05-Open_Issues.tex
\section{Open Issues \& Research Directions}

From our analysis and comparison of various BERT compression methods, we conclude that traditional model compression methods such as quantization and pruning do show benefits for BERT. Techniques specific to BERT also yield competitive results, \emph{e.g.,}~variants of knowledge distillation and methods that reduce architectural hyper-parameters. Such methods also offer insights into BERT's workings and the importance of various layers in its architecture.
However, BERT compression is still in early stages and we see multiple avenues for future research:
\begin{enumerate}[noitemsep, leftmargin=*]
    \item A very prominent feature of most BERT compression methods is their coupled nature across various encoder units, as well as its inner architecture. However, some layers might be able to handle more compression.
    Methods compressing each layer independently \cite{khetan2020schubert,tsai2020finding} have shown promising results but remain under-explored.
    
    \item The very nature of the Transformer backbone that forces the model to be parameter-heavy makes model compression for BERT more challenging. Existing work in replacing the Transformer backbone by Bi-LSTMs and CNNs has yielded extraordinary compression ratios, but with a sizable drop in accuracy. This suggests further exploration of more complicated variations of these models and hybrid Bi-LSTM/CNN/Transformer models to limit the loss in performance~\cite{tian2019waldorf}.
    
    \item Many existing methods for BERT compression only work on specific parts of the model. However, we can combine such complementary methods to achieve better overall model compression performance. We have seen in \textit{Effectiveness of the Compression Methods} that such compound compression methods perform better than their individual counterparts \cite{tambe2020edgebert,hou2020dynabert}, and thus more exploration in combining various existing methods is needed.
\end{enumerate}